\documentclass[journal]{IEEEtran}
\usepackage{graphicx}
\usepackage{booktabs}
\usepackage{subfigure}
\usepackage{overpic}
\usepackage{amsmath}
\usepackage{flushend}
\usepackage{amssymb}
\usepackage{multirow}
\usepackage{makecell}
\usepackage{fontawesome}
\usepackage{color, soul}
\usepackage{hyperref}
\hypersetup{
	colorlinks=true,
	linkcolor=blue,
	urlcolor=blue,
	citecolor=blue}
\usepackage{colortbl}
\usepackage[dvipsnames]{xcolor}
\definecolor{bg}{HTML}{e0f1ff}

\begin{document}

\title{Axial-Relation Guided Fusion State Space Model for Optical-Elevation Sensing Image Segmentation}

\author{
    Feng Gao, \emph{Member, IEEE},
    Zhilin Jin, 
    Yanhai Gan, 
    Junyu Dong, \emph{Member, IEEE},  \\
    and Qian Du, \emph{Fellow, IEEE}
\thanks{This work was supported in part by Key R \& D Program of Shandong Province under Grant 2025CXPT185, and in part by the Natural Science Foundation of Shandong Province under Grant ZR2024MF020. (\textit{Corresponding author: Yanhai Gan})

Feng Gao, Zhilin Jin, Yanhai Gan and Junyu Dong are with the State Key Laboratory of Physical Oceanography, Ocean University of China, Qingdao 266100, China. (email: ganyanhai@ouc.edu.cn)

Qian Du is with the Department of Electrical and Computer Engineering, Mississippi State University, Starkville, MS 39762 USA.}}

\markboth{IEEE GEOSCIENCE AND REMOTE SENSING LETTERS}{Shell}

\maketitle

\begin{abstract}
Semantic segmentation of multi-source remote sensing images is a fundamental task for Earth observation applications. Existing methods often struggle with insufficient multi-scale context modeling and suboptimal cross-modal feature fusion, limiting their performance in complex high-resolution scenes. To this end, we propose Axial-Relation Guided Fusion Mamba (ARG-Mamba), a state space model–based framework for optical-elevation remote sensing image segmentation. Specifically, we introduce a Multi-Scale State Space Module to capture both fine-grained local details and global contextual dependencies with linear computational complexity. Moreover, an Axial-Relation Guided Fusion Module is designed to explicitly model global cross-modal correlations along horizontal and vertical axes, enabling efficient feature fusion between optical and elevation modalities. Extensive experiments conducted on the ISPRS Vaihingen and Potsdam datasets demonstrate that our ARG-Mamba consistently outperforms state-of-the-art methods while maintaining favorable computational efficiency. The code will be made publicly available at \url{https://github.com/oucailab/ARG-Mamba}.

\end{abstract}


\IEEEpeerreviewmaketitle

\section{Introduction}

\IEEEPARstart{S}{emantic} segmentation of high-resolution remote sensing imagery plays a critical role in numerous Earth observation applications, such as urban planning, land cover mapping, and disaster management \cite{ma2019deep}. Most existing semantic segmentation methods are primarily developed based on optical remote sensing images, which mainly capture spectral and textural information of surface materials \cite{afenet25}. However, relying solely on optical imagery often leads to limited segmentation performance in complex scenes. In contrast, elevation data from digital surface model (DSM) provide complementary geometric and structural cues, which are highly discriminative for distinguishing land-cover categories \cite{fjy24grsl}. Therefore, in this letter, we mainly focus on optical-elevation remote sensing image segmentation.

\begin{figure*}[!t]
    \centering
    \includegraphics[width=0.75\textwidth]{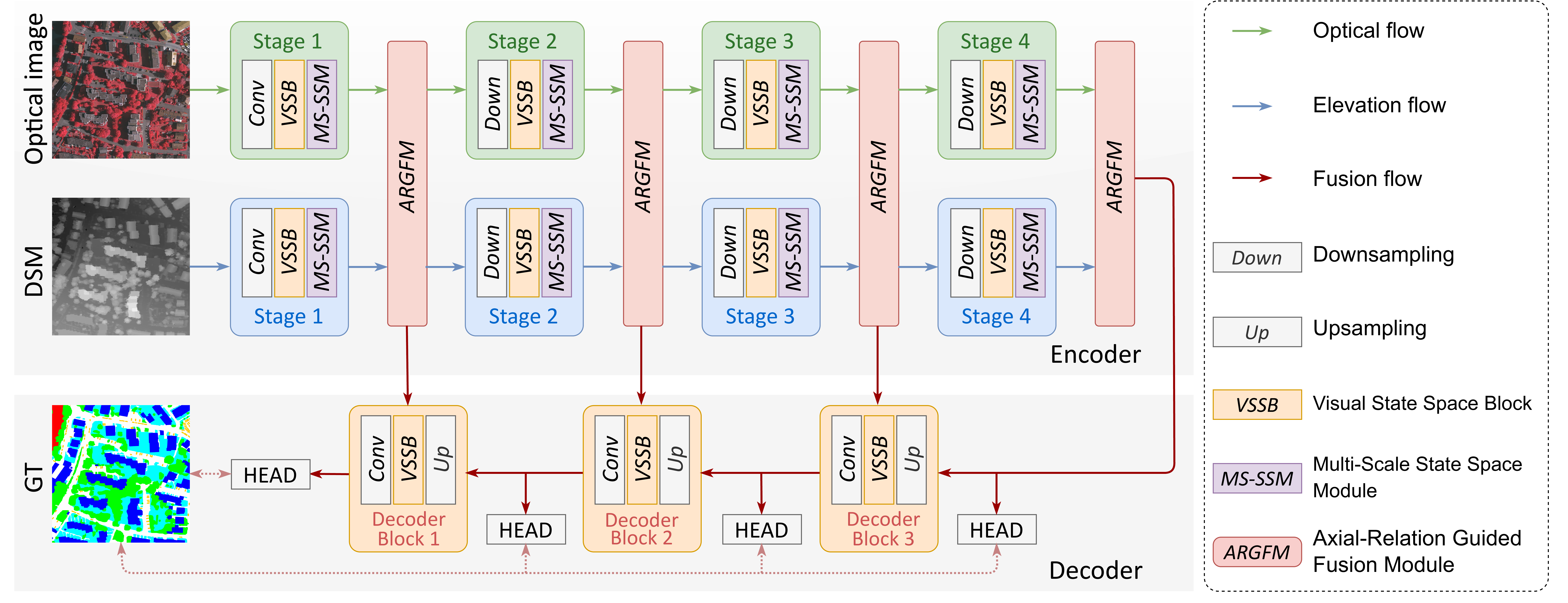} 
    \caption{The overall architecture of \textbf{ARG-Mamba}. It is a dual-stream encoder–decoder framework for optical-elevation semantic segmentation. It processes optical and DSM inputs in parallel using hierarchical \textbf{Visual State Space Blocks (VSSBs)} and \textbf{Multi-Scale State Space Modules (MS-SSMs)} to efficiently capture long-range dependencies with linear complexity. At each encoder stage, an \textbf{Axial-Relation Guided Fusion Module (ARGFM)} explicitly models global cross-modal correlations via axial relations, enabling progressive interaction between modalities.}
    \label{fig-frame}
\end{figure*}

Numerous remote sensing image semantic segmentation models have been proposed based on convolutional neural networks (CNNs) and Transformer architectures \cite{ma2024ftransunet}. In terms of multi-modal interaction, fusion strategies vary from early concatenation \cite{Weng2022} to complex feature-level integration. Edge-guided convolution \cite{jjh23grsl}, multi-scale modeling \cite{xt23tgrs}, and attentive frequency decoupling \cite{lx25tgrs} have been employed for complex contextual feature modeling. Recently, newly emerged state space models (SSMs), particularly Mamba \cite{gu24mamba}, have demonstrated strong capability in capturing long-range dependencies with linear computational complexity. SSMs have achieved competitive performance while offering excellent efficiency advantages across a wide range of remote sensing data interpretation tasks \cite{msfmamba}.

Leveraging SSMs for optical-elevation remote sensing image segmentation is a non-trivial task due to the following two challenges: \textbf{(1) \textit{Insufficient exploitation of multi-scale spatial information.}} Remote sensing images inherently exhibit strong multi-scale characteristics. While most SSM-based methods typically operate on a single-scale token sequence, making it difficult to simultaneously capture fine-grained local details and coarse global contextual cues. \textbf{(2) \textit{Cross-modal complementarity between optical and elevation data needs to be enhanced.}} Optical images capture rich spectral and texture cues, while elevation data encode critical geometric and structural characteristics. The semantic relevance of optical and elevation features varies across object categories, requiring adaptive and context-aware fusion.

To address the abovementioned challenges, we propose \textbf{A}xial-\textbf{R}elation \textbf{G}uided Fusion \textbf{Mamba} (\textbf{ARG-Mamba}) network for optical-elevation remote sensing image segmentation. First, to enhance multi-scale feature representation, we propose the \textit{Multi-Scale State Space Module (MS-SSM)}. By performing state space modeling at different spatial scales, MS-SSM captures both long-range global dependencies and fine-grained local structures. Second, to enhance cross-modal complementarity, we present the \textit{Axial-Relation Guided Fusion Module (ARGFM)}. It  models contextual relations along horizontal and vertical axes to fuse heterogeneous features. By leveraging axial relations, ARGFM effectively captures long-range structural dependencies across modalities.

The contributions of our ARG-Mamba are threefold:

\begin{itemize}

\item We propose ARG-Mamba framework for multi-source remote sensing image segmentation, which integrates state space models with explicit axial-relation guided fusion to efficiently capture long-range dependencies.

\item MS-SSM is introduced to address the scale variation in high-resolution remote sensing images, enabling joint modeling of fine-grained local details and global contextual information in a unified framework.

\item Extensive experiments are conducted on two widely used benchmark optical-elevation datasets. The results demonstrate that our ARG-Mamba outperforms state-of-the-art methods.

\end{itemize}

\section{Methodology}

The framework of Axial-Relation Guided Fusion Mamba (ARG-Mamba) network is shown in Fig. \ref{fig-frame}. It is a dual-stream encoder-decoder network designed for optical-elevation semantic segmentation. It integrates SSM with explicit axial-relation guided fusion, enabling efficient long-range dependency modeling and effective cross-modal interaction throughout the network hierarchy. Specifically, the encoder adopts a two-branch architecture, and each branch consists of four hierarchical stages with progressively reduced spatial resolution. Each stage is composed of a Visual State Space Block (VSSB) \cite{liu24vmamba} and a Multi-Scale State Space Module (MS-SSM), which jointly capture both local context and long-range dependencies with linear computational complexity. To enable effective cross-modal interaction, an Axial-Relation Guided Fusion Module (ARGFM) is inserted between corresponding stages of the encoders. ARGFM explicitly models global cross-modal correlations by decomposing dense relations into axial (row-wise and column-wise) interactions, allowing optical and elevation features to be mutually enhanced in a structured and efficient manner. In the decoder, high-level fused features are gradually upsampled and refined through a sequence of decoder blocks. Deep supervision is applied by attaching auxiliary prediction heads at multiple decoder stages.

\begin{figure}
  \centering
  \includegraphics[width=0.65\linewidth]{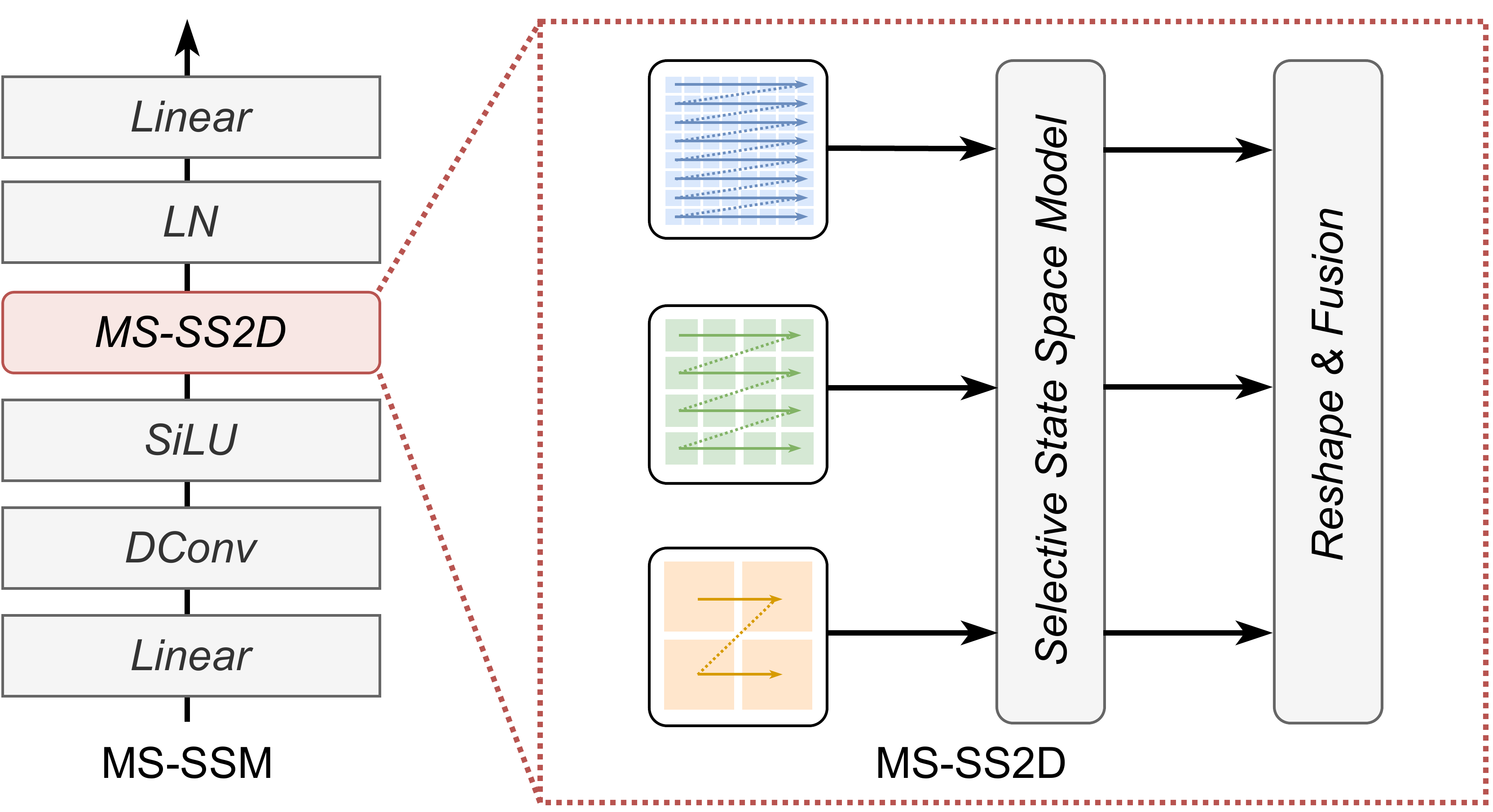}
  \caption{Illustration of the Multi-Scale State Space Module (MS-SSM).}
  \label{fig-ms-ssm}
\end{figure}

\subsection{Multi-Scale State Space Module (MS-SSM)}

Details of the MS-SSM is shown in Fig. \ref{fig-ms-ssm}. It has similar architecture of VSSB in VMamba \cite{liu24vmamba}. Meanwhile, the S6 block in Mamba is substituted with the newly proposed multi-scale selective scan (MS-SS2D). To be specific, we employ multiple scanning scales, denoted as $\{s_i\}^N_{i=1}$. Given an input feature map $x \in \mathbb{R}^{C \times H \times W}$, we split the $x$ into $N$ sub-features $\{x_i\}^N_{i=1}$ along the channel dimension. For the sub-feature $x_i$ with scan scale $s_i$, we divide $x_i$ into $\frac{H}{s_i}\times \frac{W}{s_i}$ non-overlapping windows. Then, the SSM scanning is performed within the windows, resulting in the scan input $\bar{x}_i$. The local window scan restricts scanning distance to minimize attenuation and noise between tokens. 

For all sub-features, we utilize the two-way scanning method to generate features at the specified scale. Although employing more scanning directions could potentially capture richer spatial relations, we restrict it to two-way scanning to minimize computational redundancy across multiple scales, ensuring the efficiency of MS-SSM. This method generates two token sequences. The resulting sequences are then processed by SSM \cite{gu24mamba}, and combined to produce the output feature $\bar{y} = \{ \bar{y}_i\}^N_{i=1}$ as follows:
\begin{equation}
  \bar{y}_i=\textrm{SS2D} (\bar{x}_i) \in \mathbb{R}^{C_i\times H\times W},
\end{equation}
where $C_i$ is the number of channels of the $i$-th scale. Finally, we concatenate the features from all scales along the channel dimension, yielding the final output feature. To facilitate cross-scale interaction, the concatenated feature is processed by a linear projection and a depth-wise convolution (DConv), which adaptively merges multi-scale cues and reconciles channel-wise redundancies. In our implementations, we use four scales $\{1,2,4,8\}$. This set is selected to align with the typical spatial hierarchies of remote sensing objects (e.g., small cars to large buildings), ensuring a comprehensive receptive field.

\begin{figure}
  \centering
  \includegraphics[width=0.8\linewidth]{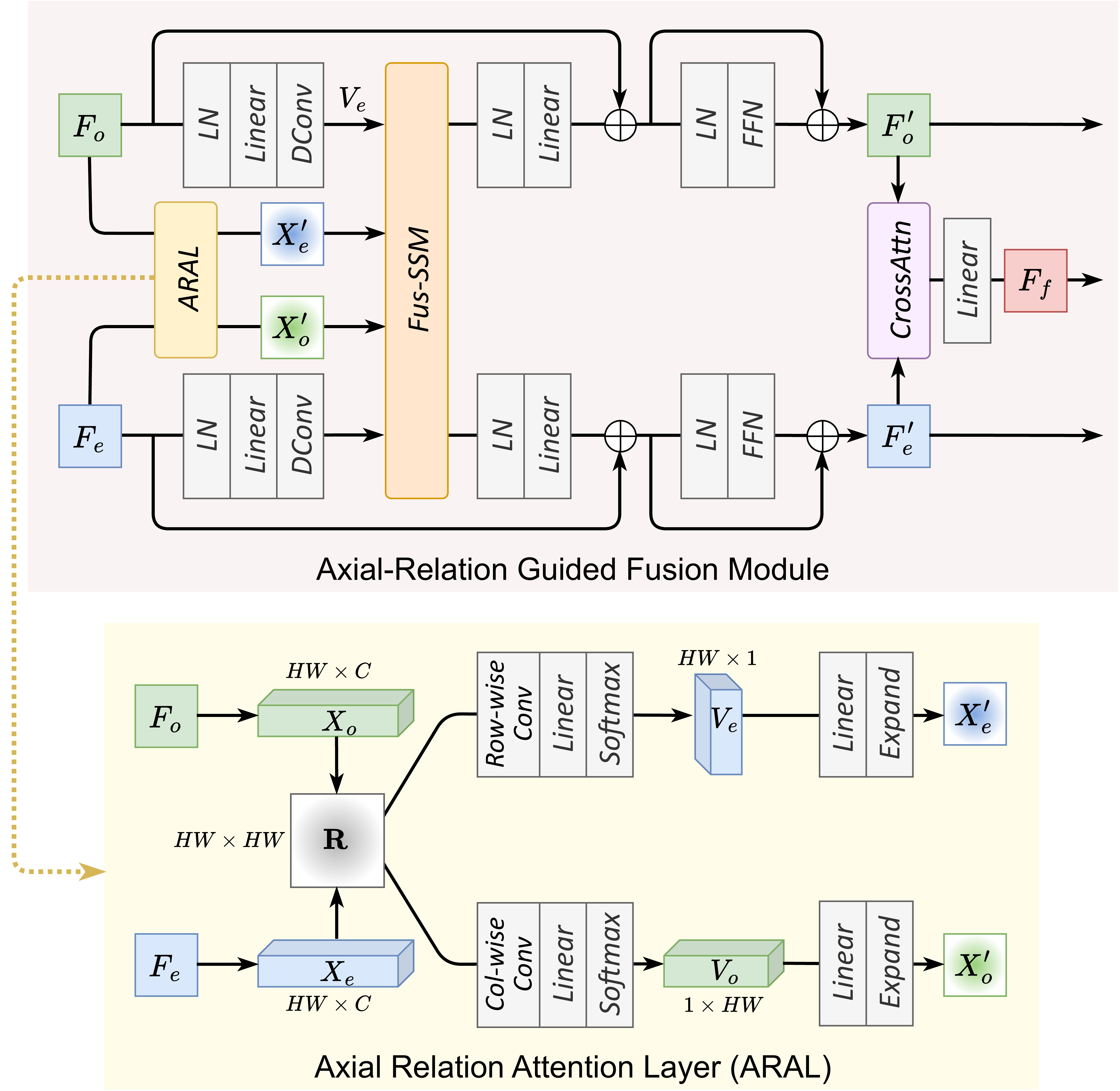}
  \caption{Illustration of the Axial-Relation Guided Fusion Module (ARGFM).}
  \label{fig-argfm}
\end{figure}

\subsection{Axial-Relation Guided Fusion Module (ARGFM)}

The Axial-Relation Guided Fusion module is designed to achieve efficient and globally-aware cross-modal fusion by jointly exploiting axial relation attention and state space sequence modeling. As shown in Fig. \ref{fig-argfm}, it takes optical and elevation features as input and produces a fused representation with enhanced cross-modal consistency and long-range dependency modeling. Specifically, the optical and elevation feature maps $F_o$ and $F_e$ are first projected using a lightweight embedding layer to generate local context enhanced features $\hat{X}_o$ and $\hat{X}_e$. Specifically, the input maps $F_o, F_e \in \mathbb{R}^{C \times H \times W}$ are reshaped and projected to the latent dimension $C$. 

To explicitly model global cross-modal dependencies, we introduce the Axial Relation Attention Layer (ARAL), which produces relation-aware modulation features:
\begin{equation}
  (X'_o, X'_e) = \mathrm{ARAL}(F_o, F_e),
\end{equation}
where $X'_o$ and $X'_e$ are the relation-enhanced optical and elevation tokens, respectively. ARAL encodes global cross-modal correlations via axial (row–column) relation decomposition, enabling efficient long-range interaction. Details of ARAL are described in Section \ref{aral}.

The projected features and relation-aware tokens are jointly fed into a fusion SSM block (Fus-SSM) \cite{msfmamba} to model long-range dependencies across modalities:
\begin{equation}
 (Z_o, Z_e) = \mathrm{Fus\text{-}SSM}(\tilde{X}_o, \tilde{X}_e, X'_o, X'_e),
\end{equation}
Specifically, $\tilde{X}_o, \tilde{X}_e \in \mathbb{R}^{HW \times C}$ denote the local context tokens, while $X_o', X_e' \in \mathbb{R}^{HW \times C}$ serve as relation-aware modulators derived from the ARAL. Within Fus-SSM, these four inputs are cross-concatenated and processed via selective scanning to facilitate deep feature interaction between spectral and geometric cues. The SSM propagates information along the token sequence with linear complexity. The SSM outputs are then linearly projected, followed by a feed-forward network (FFN) with residual connections to generate $F'_o$ and $F'_e$.

Finally, cross-attention layer and linear projection are employed to explicitly integrate the refined optical and elevation representations:
\begin{equation}
F_f = \mathrm{FC}(\mathrm{CrossAttn}(F'_o, F'_e)),					
\end{equation}
where $\mathrm{CrossAttn}$ denotes the cross-attention computation, and $\mathrm{FC}$ denotes the linear layer. $F'_o$ and $F'_e$ are forwarded to the next stages of the optical and elevation streams, respectively, while $F_f$ is passed to the corresponding decoder block.

\subsection{Axial Relation Attention Layer (ARAL)} 
\label{aral}

Details of the ARAL are shown in Fig. \ref{fig-argfm}. Our ARAL constructs a dense relation matrix between optical and elevation tokens. Instead of performing full 2D attention on the relation matrix, ARAL decomposes it into row-wise and column-wise axial attentions, which capture complementary interaction patterns along two orthogonal spatial axes while reducing computational complexity. 

Given the optical feature $F_o \in \mathbb{R}^{C \times H \times W}$ and the elevation feature $F_e \in \mathbb{R}^{C \times H \times W}$, we first reshape them into token representations along the spatial dimension as $X_o = \mathrm{reshape}(F_o) \in \mathbb{R}^{HW \times C}$, $X_e = \mathrm{reshape}(F_e) \in \mathbb{R}^{HW \times C}$. 

To explicitly capture the global cross-modal correlations between optical and elevation features, we compute a dense relation matrix as:
\begin{equation}
\mathbf{R} = X_o X_e^{\top} \in \mathbb{R}^{HW \times HW},
\end{equation}
where each element $\mathbf{R}_{ij}$ measures the interaction strength between the $i$-th optical token and the $j$-th elevation token. Although the formulation of $R$ involves a computational complexity of $O((HW)^2)$, this potential bottleneck is mitigated by the subsequent axial decomposition.

After that, we separately perform row-wise and column-wise convolution on the relation matrix $\mathbf{R}$ for modality-specific pooling. By decomposing the dense 2D relations into row-wise and column-wise interactions, the computational complexity is significantly reduced from $O((HW)^2)$ to $O(HW(H+W))$. This orthogonal decomposition is specifically chosen to align with the grid-like structural layout of remote sensing objects, such as road networks and building alignments, while maintaining computational parsimony. The obtained features are processed by linear layer and Softmax, respectively. Then, two weighted vectors are computed as follows:

\begin{equation}
  V_e = \mathrm{Softmax}(\mathrm{FC}(\mathrm{Conv}_{row}(\mathbf{R}))) \in \mathbb{R}^{HW\times1},
\end{equation}
\begin{equation}
  V_o = \mathrm{Softmax}(\mathrm{FC}(\mathrm{Conv}_{col}(\mathbf{R}))) \in \mathbb{R}^{1\times HW},
\end{equation}
where $\mathrm{Conv}_{row}$ and $\mathrm{Conv}_{col}$  reduce the relation matrix $R \in \mathbb{R}^{HW \times HW}$ to $HW \times 1$ and $1 \times HW$, respectively. $V_e$ and $V_o$ are expanded back to $\mathbb{R}^{HW \times C}$ to produce the relation-aware tokens $X'_e$, $X'_o$.

\section{Experimental Results and Analysis}

\subsection{Datasets and Experimental Setting}

We evaluate the performance of the proposed ARG-Mamba on two multi-source remote sensing datasets for semantic segmentation. Specifically, the first dataset is the ISPRS Vaihingen dataset. It consists of high-resolution optical and DSM data with the Ground Sampling Distance (GSD) of 9 cm. The second dataset is the ISPRS Potsdam dataset, which consists of optical and DSM data with the GSD of 5 cm. Both datasets contain 6 classes segmentation. The original images are cropped into $512 \times 512$ patches for training and testing. Our model was implemented using PyTorch and trained on a single NVIDIA RTX 4090 GPU. We used the Adam optimizer and the training phase spanned 200 epochs. The initial learning rate is set to 0.0001, adjusted by a cosine annealing strategy. 

To evaluate the performance of the proposed ARG-Mamba, we compare it against six state-of-the-art methods: PSTNet \cite{shivakumar2020pst900}, FEANet \cite{deng2021feanet}, GMNet \cite{zhou2021gmnet}, CMNeXt \cite{zhang2023cmnext}, FTransUNet \cite{ma2024ftransunet}, and MFNet \cite{ma2025mfnet}. These methods are assessed through quantitative metrics, including Overall Accuracy (OA), Mean Intersection over Union (mIoU), and F1-score (F1). OA represents the proportion of correctly classified pixels relative to the total number of pixels in the dataset. mIoU calculates the average ratio of the intersection to the union of the predicted and ground truth segments for all classes. The F1-score offers a comprehensive evaluation by calculating the harmonic mean of precision and recall.

\subsection{Experimental Results and Discussion}

\textbf{Quantitative Analysis.} Tables \ref{tab:vaihingen_res} and \ref{tab:potsdam_res} report quantitative comparisons on the ISPRS Vaihingen and Potsdam datasets, respectively. Overall, ARG-Mamba consistently achieves the best performance across all three global metrics (F1, OA, and mIoU) on both datasets, demonstrating its strong segmentation capability. On the ISPRS Vaihingen dataset, ARG-Mamba obtains the highest mIoU, outperforming all competing methods. In terms of per-class performance, ARG-Mamba achieves the best results on Impervious Surface, Low Vegetation, Tree, and Car, with particularly notable improvements on the Car class, indicating its advantage in recognizing small and complex objects. These results highlight the effectiveness of ARG-Mamba in modeling both global context and fine-grained structures. On the ISPRS Potsdam dataset, ARG-Mamba again delivers the best overall performance. It also achieves the best accuracy on Impervious Surface and Low Vegetation, and remains highly competitive on Building, Tree, and Car classes. Compared with strong Transformer- and CNN-based baselines, ARG-Mamba shows more balanced and robust performance across different land-cover categories.

\begin{figure*}
  \centering
  \includegraphics[width=0.75\linewidth]{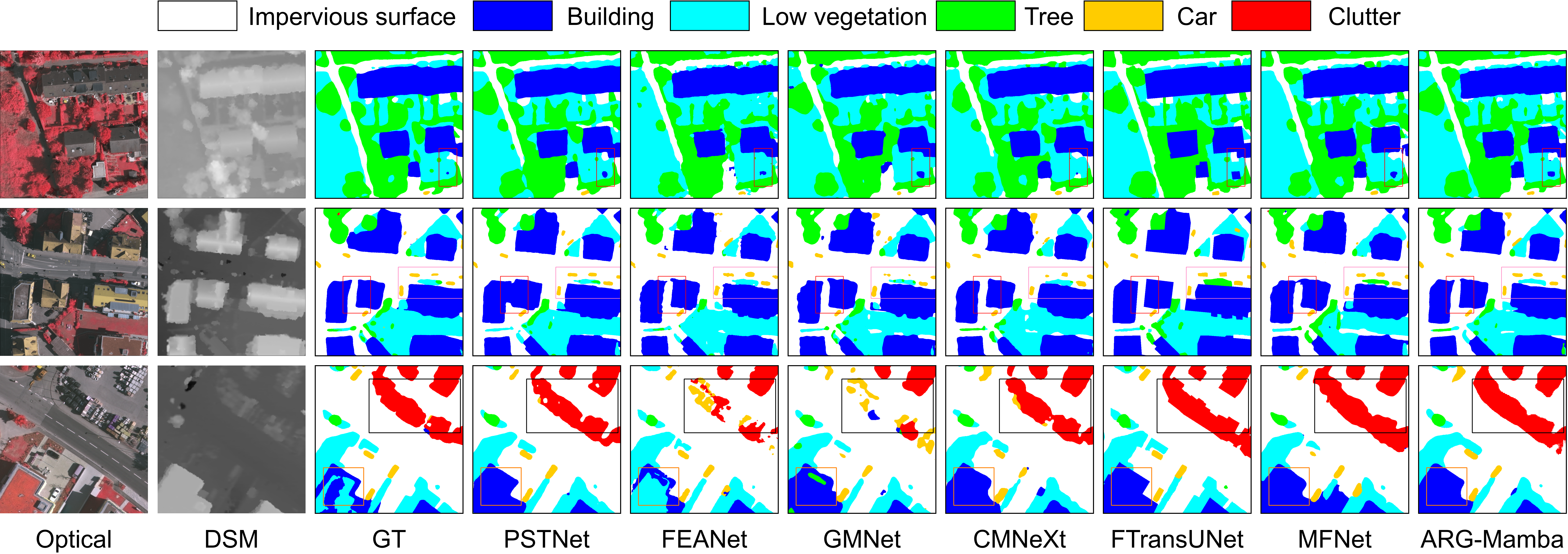}
  \caption{Classification results of different methods on the ISPRS Vaihingen dataset.}
  \label{fig-res}
\end{figure*}

\begin{figure}
  \centering
  \includegraphics[width=0.9\linewidth]{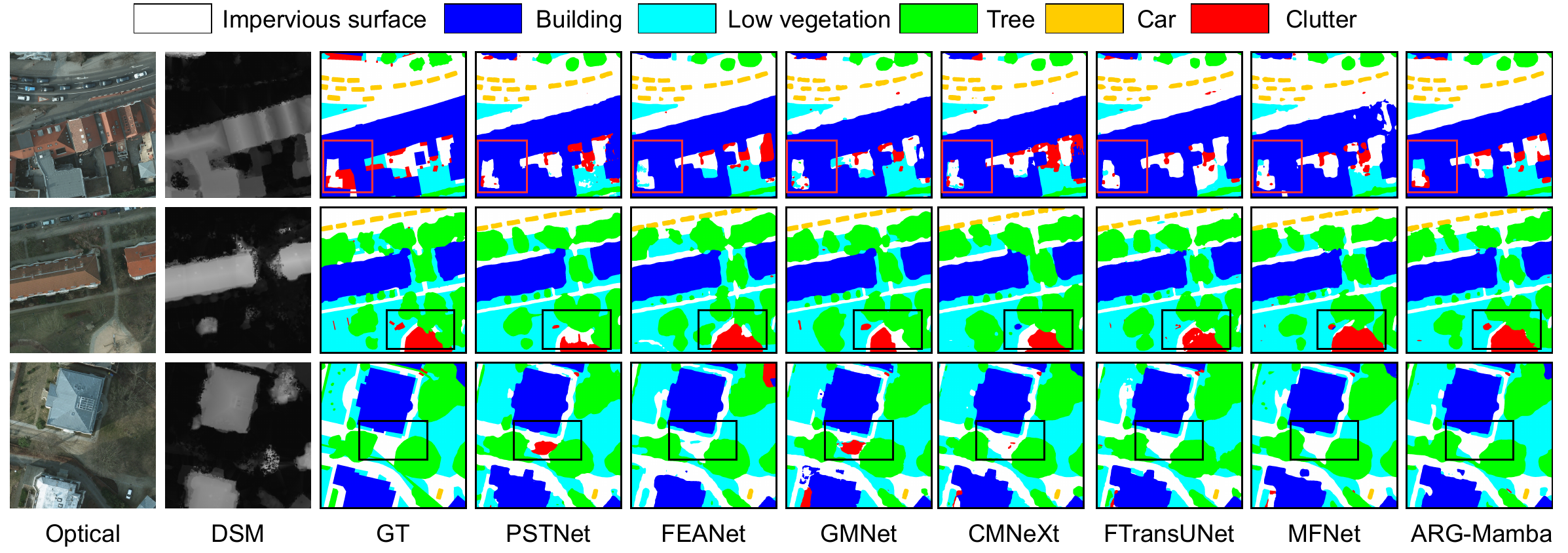}
  \caption{Classification results of different methods on the ISPRS Potsdam dataset.}
  \label{fig-pots}
\end{figure}

\begin{table}[]
\centering
\caption{Experimental results on the ISPRS Vaihingen dataset. The \textbf{bold} and \underline{underline} denote the best and second results.}
\setlength{\tabcolsep}{4pt} 
\resizebox{\columnwidth}{!}{
\begin{tabular}{l|ccc|ccccc}
\toprule
Method & F1 & OA & mIoU & Imp. & Building & LowVeg. & Tree & Car \\
\midrule
PSTNet \cite{shivakumar2020pst900} & 88.50 & 92.23 & 79.99 & 92.51 & 90.62 & 70.83 & 80.01 & 65.99 \\
FEANet \cite{deng2021feanet} & 86.64 & 91.97 & 77.50 & 92.45 & 90.27 & 69.91 & 79.11 & 55.75 \\
GMNet \cite{zhou2021gmnet} & 87.42 & 92.32 & 78.63 & 92.15 & 91.52 & 70.57 & 80.73 & 58.16 \\
CMNeXt \cite{zhang2023cmnext} & 89.61 & \underline{92.72} & 81.61 & 93.16 & 90.23 & \underline{72.27} & 80.62 & 71.75 \\
FTransUNet \cite{ma2024ftransunet} & 88.73 & 92.69 & 80.42 & \underline{93.27} & \textbf{92.45} & 70.75 & 79.78 & 65.87 \\
MFNet \cite{ma2025mfnet} & \underline{90.10} & 92.94 & \underline{82.42} & 93.26 & \underline{92.11} & 72.11 & \underline{80.89} & \underline{73.73} \\
\rowcolor{bg}\textbf{ARG-Mamba} & \textbf{90.96} & \textbf{93.28} & \textbf{83.72} & \textbf{93.82} & 90.49 & \textbf{74.37} & \textbf{81.99} & \textbf{77.94} \\
\bottomrule
\end{tabular}}
\label{tab:vaihingen_res}
\end{table}

\begin{table}[]
\centering
\caption{Experimental results on the ISPRS Potsdam dataset. The \textbf{bold} and \underline{underline} denote the best and second results.}
\setlength{\tabcolsep}{4pt} 
\resizebox{\columnwidth}{!}{
\begin{tabular}{l|ccc|ccccc}
\toprule
Method & F1 & OA & mIoU & Imp. & Building & LowVeg. & Tree & Car \\
\midrule
PSTNet \cite{shivakumar2020pst900} & 89.55 & 88.86 & 81.47 & 85.09 & 94.05 & 72.26 & 72.17 & 83.75 \\
FEANet \cite{deng2021feanet}  & 91.02 & 90.26 & 83.82 & 87.32 & \textbf{94.52} & 75.62 & 74.86 & 86.79 \\
GMNet \cite{zhou2021gmnet} & 90.44 & 89.66 & 82.88 & 86.55 & 93.77 & 73.33 & 74.86 & 85.88 \\
CMNeXt \cite{zhang2023cmnext} & 91.28 & 90.09 & 84.21 & 86.35 & 92.98 & \underline{76.54} & 76.07 & 89.12 \\
FTransUNet \cite{ma2024ftransunet}  & 90.93 & 89.79 & 83.69 & \underline{87.40} & 93.23 & 74.74 & 74.61 & 88.43 \\
MFNet \cite{ma2025mfnet} & \underline{91.80} & \underline{90.75} & \underline{85.10} & 87.20 & 93.90 & 76.50 & \underline{77.20} & \underline{89.15} \\
\rowcolor{bg}\textbf{ARG-Mamba} & \textbf{91.93} & \textbf{90.89} & \textbf{85.30} & \textbf{87.44} & \underline{94.14} & \textbf{78.09} & \textbf{77.37} & \textbf{89.46} \\
\bottomrule
\end{tabular}}
\label{tab:potsdam_res}
\end{table}

\textbf{Qualitative Analysis.} Fig. \ref{fig-res} and Fig. \ref{fig-pots} present qualitative segmentation results of different methods on the ISPRS Vaihingen and Potsdam datasets, respectively. Compared with CNN-based and Transformer-based baselines, ARG-Mamba produces segmentation maps that are visually closer to the ground truth on both datasets, exhibiting clearer object boundaries and more consistent region labeling. Specifically, ARG-Mamba delivers sharper boundaries in building extraction and preserves fine-grained structures in vegetation classes better than competitors. It also shows a distinct advantage in recognizing small objects like cars, where other methods often fail. The reduced clutter noise and improved spatial consistency further indicate that ARG-Mamba benefits from its enhanced multi-scale context modeling capability.

\textbf{Computational Efficiency.} Table \ref{tab:comp} summarizes the computational costs of different methods in terms of model parameters, FLOPs, and inference time. Compared with Transformer-based and hybrid architectures, ARG-Mamba requires significantly fewer parameters and lower FLOPs than FEANet, GMNet, FTransUNet, and MFNet, while also achieving faster inference speed. In particular, MFNet and FEANet exhibit substantially higher model complexity, leading to increased computational overhead. Although PSTNet has the lowest computational cost, its segmentation accuracy is notably inferior to our ARG-Mamba, as shown in Tables \ref{tab:vaihingen_res} and \ref{tab:potsdam_res}. Therefore, the results demonstrate that our ARG-Mamba offers an efficient and lightweight design without sacrificing accuracy, making it well suited for large-scale remote sensing image segmentation tasks.

\begin{table}[htbp]
  \centering
  \caption{Computational costs of different methods.}
  \label{tab:comp}
  \resizebox{2.5in}{!}{
  \begin{tabular}{l|ccc}
    \toprule
    Method & Params (M) & FLOPs (G) & Time (s) \\
    \midrule
    PSTNet \cite{shivakumar2020pst900} & 20.4 & 15.2 & 0.185 \\
    FEANet \cite{deng2021feanet} & 255.3 & 145.8 & 0.528 \\
    GMNet \cite{zhou2021gmnet} & 153.4 & 89.6 & 0.356 \\
    CMNeXt \cite{zhang2023cmnext} & 58.7 & 42.5 & 0.242 \\
    FTransUNet \cite{ma2024ftransunet} & 179.6 & 112.3 & 0.412 \\
    MFNet \cite{ma2025mfnet} & 313.2 & 180.2 & 0.589 \\
    \rowcolor{bg} ARG-Mamba & 77.1 & 56.4 & 0.268 \\
    \bottomrule
  \end{tabular}}
\end{table}

\begin{table}[htbp]
  \centering
  \caption{Influence of MS-SSM and ARGFM on the segmentation results. The \textbf{bold} denote the best results.}
  \label{tab:ablation}
  \resizebox{2in}{!}{
  \begin{tabular}{cc|cc}
  \toprule
  MS-SSM & ARGFM & Vaihingen & Potsdam \\
    \midrule
    & & 80.15 & 81.08 \\
    & \checkmark & 82.12 & 83.18 \\
  \checkmark  &  & 82.45 & 83.60 \\
\rowcolor{bg}  \checkmark  & \checkmark & 83.72 & 85.30 \\
    \bottomrule
  \end{tabular}}
\end{table}

\subsection{Ablation Study}

Table \ref{tab:ablation} shows the impact of MS-SSM and ARGFM on the segmentation performance in terms of mIoU on both datasets. Using either MS-SSM or ARGFM alone improves the mIoU on both Vaihingen and Potsdam datasets compared with the baseline. When both modules are combined, the model achieves the best results, reaching 83.72\% on Vaihingen and 85.30\% on Potsdam, demonstrating their complementary effectiveness. Additionally, we investigated the sensitivity of key hyperparameters. Results indicate that varying the number of scales $N$ from 2 to 6 yields marginal fluctuations in mIoU, and increasing the axial convolution kernel size beyond $3 \times 3$ provides diminishing returns while increasing latency. These findings confirm that the current configuration ($N=4$ and $3 \times 3$ kernels) strikes an optimal balance between accuracy and efficiency.

\section{Conclusion}

In this letter, we presented ARG-Mamba, a state space model-based framework for optical-elevation remote sensing image segmentation. To exploit the multi-scale spatial information, we proposed MS-SSM to effectively model both fine-grained local details and long-range global context across multiple spatial scales with linear computational complexity. In addition, we proposed ARGFM to explicitly model cross-modal correlations along spatial axes, facilitating efficient fusion of cross-modal features. Extensive experiments on the ISPRS Vaihingen and Potsdam datasets demonstrate that our ARG-Mamba consistently outperforms state-of-the-art methods in both accuracy and efficiency.

\bibliographystyle{IEEEtran}
\bibliography{re}

\end{document}